\documentclass[preprint,5p]{elsarticle}

\usepackage{setspace}
\usepackage{xcolor}
\usepackage{multirow}
\usepackage{color}

\usepackage{amsmath}
\usepackage{amsthm}
\usepackage{amsbsy}
\usepackage{amssymb}
\usepackage{dsfont}
\usepackage{nicefrac}
\usepackage{siunitx}
\usepackage{etoolbox}
\usepackage{xcolor}
\usepackage{epigraph}
\usepackage{arydshln}
\usepackage[first=0, last=9]{lcg}
\usepackage{booktabs}
\usepackage[normalem]{ulem}
\usepackage{lipsum}

\renewcommand{\bfseries}{\fontseries{b}\selectfont}
\newrobustcmd{\B}{\bfseries}

\usepackage{scalerel}
\usepackage{tikz}
\usetikzlibrary{svg.path}

\definecolor{orcidlogocol}{HTML}{A6CE39}
\tikzset{
    orcidlogo/.pic={
        \fill[orcidlogocol] svg{M256,128c0,70.7-57.3,128-128,128C57.3,256,0,198.7,0,128C0,57.3,57.3,0,128,0C198.7,0,256,57.3,256,128z};
        \fill[white] svg{M86.3,186.2H70.9V79.1h15.4v48.4V186.2z}
        svg{M108.9,79.1h41.6c39.6,0,57,28.3,57,53.6c0,27.5-21.5,53.6-56.8,53.6h-41.8V79.1z M124.3,172.4h24.5c34.9,0,42.9-26.5,42.9-39.7c0-21.5-13.7-39.7-43.7-39.7h-23.7V172.4z}
        svg{M88.7,56.8c0,5.5-4.5,10.1-10.1,10.1c-5.6,0-10.1-4.6-10.1-10.1c0-5.6,4.5-10.1,10.1-10.1C84.2,46.7,88.7,51.3,88.7,56.8z};
    }
}

\newcommand\orcidicon[1]{\href{https://orcid.org/#1}{\mbox{\scalerel*{
\begin{tikzpicture}[yscale=-1,transform shape]
\pic{orcidlogo};
\end{tikzpicture}
}{|}}}}

\usepackage{hyperref}% <--- load after everything 
\usepackage{enumitem}

\def \ie {\emph{i.e.},}
\def \eg {\emph{e.g.},}
\def \etal {\emph{et al.} }

\newcommand{\dataname}{$\text{Font}^2$}
\newcommand{\numfonts}{10400 }
\newcommand{\numwords}{10400 }
\newcommand{\numsamples}{108160000}

\newcommand{\tit}[1]{\smallbreak\noindent\textbf{#1.}}

\begin{document}

\begin{frontmatter}

\title{Evaluating Synthetic Pre-Training for Handwriting Processing Tasks}

\author{Vittorio~Pippi\corref{cor1}}
\cortext[cor1]{Corresponding author}
\ead{vittorio.pippi@unimore.it}

\author{Silvia~Cascianelli}
\ead{silvia.cascianelli@unimore.it}

\author{Lorenzo~Baraldi}
\ead{lorenzo.baraldi@unimore.it}

\author{Rita~Cucchiara}
\ead{rita.cucchiara@unimore.it}

\address{University of Modena and Reggio Emilia, Via Vivarelli 10, Modena, Italy}

\begin{abstract}
In this work, we explore massive pre-training on synthetic word images for enhancing the performance on four benchmark downstream handwriting analysis tasks. To this end, we build a large synthetic dataset of word images rendered in several handwriting fonts, which offers a complete supervision signal. We use it to train a simple convolutional neural network (ConvNet) with a fully supervised objective. 
The vector representations of the images obtained from the pre-trained ConvNet can then be considered as encodings of the handwriting style. We exploit such representations for Writer Retrieval, Writer Identification, Writer Verification, and Writer Classification and demonstrate that our pre-training strategy allows extracting rich representations of the writers' style that enable the aforementioned tasks with competitive results with respect to task-specific State-of-the-Art approaches.
\end{abstract}


\end{frontmatter}

%\linenumbers

\section{Introduction}
Large-scale pre-training on a large quantity of data and then applying or adapting the obtained model for downstream tasks is a common practice in several deep learning applications. This strategy is rewarding also when working on text images. Indeed, some research effort has been recently dedicated to learning robust representations of text images to be then used for text recognition~\cite{kang2020distilling,aberdam2021sequence,luo2022siman}. In this work, we build a representation of the handwriting style of a small piece of text (word images) and focus on different writer-centric tasks of handwriting processing. 

Handwriting processing tasks are related to the analysis of the handwriting style for extracting useful information about writers and document collections~\cite{atanasiu2011writer}. Such information can be exploited in different domains ranging from biometrics, forensics, digital humanities, and paleography. In particular, we consider the following tasks. 
\emph{Writer Identification}, which consists in determining the author of a piece of text out of a set of known writers. 
\emph{Writer Retrieval}, \ie~gathering from a database all the documents handwritten with a similar style as a query document. 
\emph{Writer Verification}, which is the task of determining whether two pieces of text have been handwritten by the same writer. Note that it can be performed at word level, as a signature verification task, or at document level. 
\emph{Writer Classification}, which entails automatically grouping the pieces of text in a collection based on their writer. 
These tasks have been considered also for their role in different practical applications. For example, identifying the author of a document can help automatize digital archives organization; document retrieval can be exploited for automatizing and enhancing archives consultation; 
signature verification can be used as an authentication procedure; authorship authentication is important in forensic and historical document analysis; handwriting classification can enrich the analysis of historical documents by allowing distinguishing between different writers that possibly contributed to a manuscript.

For the aforementioned tasks, the information on the handwriting style is crucial. However, extracting this information entails careful task-specific feature engineering or exploiting a learning model, which is usually data-angry and thus expensive to train with sufficient data. 
Motivated by these considerations, to obtain a robust and informative representation of the handwriting style from images of handwritten words, we devise a supervised pre-training protocol on carefully designed synthetic data. The obtained model is then applied directly to real images, and the resulting representations are input to distance-based strategies for a number of downstream tasks that focus on writer-related information. 
Note that, by working with synthetic data, we have access to complete ground truth information for a large number of samples, different from what is usually the case for pre-training large vision and vision-and-language representation models~\cite{radford2021learning,alayracflamingo}. In fact, for those models, gathering a massive amount of training data with precise ground truth information is prohibitively costly, and thus, such models are usually trained by following a weakly-supervised or self-supervised paradigm. Instead, by resorting to a carefully designed synthetic dataset, its ground truth information can be fully exploited in a supervised training setting to obtain our style encoding network.

In summary, the main goal of this work is to devise a pre-training pipeline exploiting synthetic data to obtain a single strong representation model for text images that can be successfully used for a number of downstream tasks related to handwriting analysis. Therefore, the effectiveness of the proposed strategy is extensively evaluated under different setups and in comparison with task-specific State-of-the-Art approaches. In particular, we consider training on different-sized synthetic datasets (with a different number of training categories) to explore whether the use of more training data is beneficial also in our considered setting. Experimental analysis conducted on different benchmark datasets demonstrates the robustness and discriminative power of the handwriting style features obtained with our approach.
To the best of our knowledge, this is the first work exploring synthetic pre-training on text images for a variety of downstream handwriting analysis tasks.

The remainder of this paper is organized as follows. In Section~\ref{sec:related} we give an overview of pre-training strategies for text images and of the main task-specific approaches to the downstream tasks we consider. In Section~\ref{sec:approach} we describe our proposed pre-training protocol and explain our distance-based pipelines to tackle each of the downstream tasks. In Section~\ref{sec:experiments} we report and discuss the experimental evaluation results obtained on benchmark datasets of real images, and Section~\ref{sec:conclusion} concludes the paper.

\section{Related Works}\label{sec:related}
Pre-training on a large amount of synthetic data has become a popular strategy when dealing with text images, both typewritten~\cite{gupta2016synthetic} and handwritten~\cite{kang2020pay,cascianelli2021learning}. Such pre-training phase is useful to learn a robust representation of the text images~\cite{aberdam2021sequence}, which can then be used for solving downstream tasks, such as text recognition~\cite{kang2020distilling}, detection~\cite{song2022vision}, and image text editing~\cite{luo2022siman}. 
Nonetheless, the development and application of a single model for image representation, pre-trained on a massive amount of visual and textual data, has been explored mostly for tasks on natural images~\cite{radford2021learning,alayracflamingo}, while little to no attention has been dedicated to devising a similar model for text images. In light of this, this work focuses on analyzing the feasibility of learning and using a single pre-trained handwriting style representation model to be exploited for different handwriting processing tasks such as writer identification, retrieval, and verification without the task-specific pre-processing and post-processing steps that these tasks usually require. 
In the following, we review the main approaches to these tasks.

\tit{Writer Identification} The Writer Identification task can be performed both at document level~\cite{tang2016text,christlein2018encoding,chahi2019effective,chahi2020cross,javidi2020deep,kumar2020segmentation} and word-level~\cite{brink2012writer,siddiqi2010text,he2017beyond,he2019deep,khan2019dissimilarity,he2020fragnet,he2021gr}, which is considered a more challenging scenario. 
Document-level approaches exploit either histograms of local handcrafted descriptors~\cite{brink2012writer,siddiqi2010text,he2017beyond,chahi2019effective,chahi2020cross}, or combinations of features extracted with a ConvNet on keypoints or patches~\cite{christlein2018encoding,javidi2020deep,kumar2020segmentation} or on the whole page image~\cite{tang2016text} as input to a classifier. 
To tackle the word-level scenario, State-of-the-Art solutions to enhance the performance entail multi-grain representation of the word images (both the whole image and patches~\cite{he2020fragnet,he2021gr}) and multi-task learning (with auxiliary tasks concerning the semantic content of the word image, the length, and the attributes of the characters)~\cite{he2019deep}.
Here, we consider the challenging word-level variant of the task and exploit only our style representation in a simple pipeline that does not entail relying on any auxiliary task.

\tit{Writer Retrieval} Writer Retrieval is usually performed at document level by directly comparing either aggregated vectors obtained from locally extracted handcrafted descriptors~\cite{fiel2012writer,bouibed2020score} or ConvNet features~\cite{fiel2015writer}. To enhance the performance, the retrieval results can be refined in a re-ranking phase~\cite{jordan2020re,rasoulzadeh2022writer}. The comparison between query and database vectors can be performed via standard similarity and distance metrics, such as the cosine similarity, the euclidean distance, or the $\chi^2$-distance. Alternatively, similarity and dissimilarity learning can be exploited to obtain a task-specific score for comparison~\cite{keglevic2018learning,bouibed2020multiple}. 
In this work, we perform retrieval by leveraging our style representation and standard distance metrics. Moreover, we do not apply any refinement step to better evaluate the suitability of our style representation on a number of different downstream tasks.

\tit{Writer Verification} The Writer verification task can be performed on textual documents to authenticate the identity of their author~\cite{bensefia2016writer,christlein2014writer,aubin2018off}, based on grapheme-related features~\cite{aubin2018off}, on the edit distance between handwritten words~\cite{bensefia2016writer}, or on the direct comparison of global representations of the document images~\cite{christlein2014writer}. In this work, we perform writer verification at document level based on our handwriting style representation approach. 

Another popular variant of this task is signature verification, which can be either writer-dependent~\cite{hafemann2017learning}, when relying on a collection of known genuine signatures, or writer-independent, when such a collection is not available. The latter is the setting considered also in this work. Writer-independent approaches can be based on Siamese networks~\cite{dey2017signet,avola2021r} to directly compare pairs of signatures or learn robust features that can be exploited also in a writer-dependent setting. A similar strategy entails performing metric learning to discriminate genuine from forged signatures~\cite{zhu2020point,liu2021offline,wan2021learning}. Alternatively, the task can be performed by directly evaluating standard distance metrics between pairs of ConvNet-based robust representations of the signature images~\cite{wei2019inverse,engin2020offline,li2021avn,manna2022swis}, which is also the approach we follow in this work.
Note that, in both variants, we adopt the simple pipeline based on the comparison of the images representations since it allows to better highlight the robustness of our style encoding.

\begin{figure}[t]
    \centering
    \includegraphics[width=\linewidth]{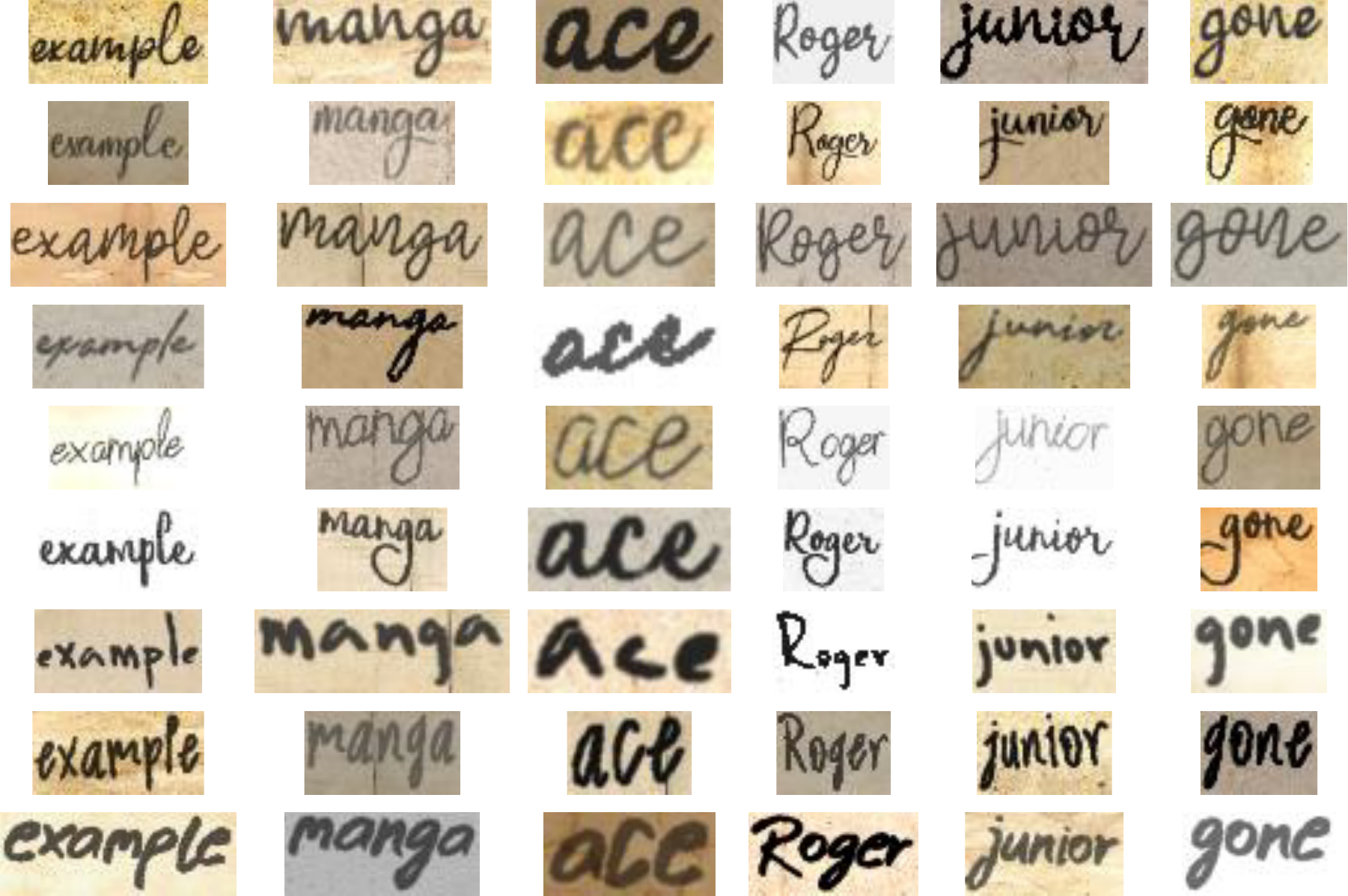}
    \caption{Exemplar images from the synthetic \dataname~dataset. Each image is labeled with the identifier of the calligraphic font used to render it.}
    \label{fig:font2}
\end{figure}

\section{Proposed Approach}\label{sec:approach}
In this work, we propose to obtain a handwriting Style Encoder via large-scale pre-training on carefully designed synthetic data. To this end, we build a synthetic dataset of word images, which we call \dataname. 

First, we scrape $S=\numfonts$ calligraphic fonts freely available online from dedicated websites, which we manually inspected to discard those in small caps and those containing decorations such as hearts or stars instead of dots or circles.
Then, we randomly choose $W=\numwords$ words of varying length from the English vocabulary. 
An alternative strategy would be using the $\numwords$ most common English words. In this work, we opt for the random choice to ensure that no letters combination (\eg~bigrams and trigrams) is penalized based on its frequency of appearance in modern English. Nevertheless, as our goal is to obtain a content-independent handwriting style representation network, the choice of the words used to build the dataset can be considered less critical to obtain such a representation. 
Finally, we render all the possible combinations between the selected fonts and words, thus obtaining a total of $\numfonts \times \numwords = \numsamples$ samples.
Such a large dataset allows us to feed the network with almost always unseen samples, mitigating the downside effect of working with fonts that, for their nature, have a very low variance between the different images. 

Note that if the synthetic data are sufficiently similar to real ones and capture the variability and characteristics of the domain of interest, the resulting pre-trained model will extract more representative features. 
In the sight of this consideration, for generating each sample, we apply the following pipeline to obtain realistic and sufficiently varied data. First, we render a word with a specific font to a white canvas. Then, we apply a random rotation and a random elastic deformation via the Thin Plate Spline transformation to introduce variance between words written with the same font. Further, we apply random gaussian blur to avoid sharp borders and simulate handwriting strokes. Moreover, we randomly select a paper-like background from a pool of images and superimpose it to the word image. To introduce further variability, we also apply random grayscale dilation and color jitter. 
Some samples from the \dataname~dataset obtained with the described procedure are reported in Fig.~\ref{fig:font2}. The code we use to generate the \dataname~dataset is available online\footnote{\url{https://github.com/aimagelab/font_square}} and can be used to automatically generate a similar synthetic dataset starting from a collection of backgrounds and fonts, possibly more specialized to the domain of interest (\eg~typewritten text or historical documents).

Dealing with synthetic data has the advantage of offering a complete supervision signal for a large number of samples. Thus, we use the \dataname~dataset to train a ResNet-18 in a supervised fashion. We chose this network as our backbone following other literature approaches dealing with text images, \eg~\cite{javidi2020deep,zhu2020point,manna2022swis}, which use it as style-encoder for its simplicity and relatively limited amount of parameters.
The network is trained to classify the image words based on the font in which they are written, regardless of the textual content, \ie~the images are associated with labels reflecting the identifier of the font used to render them. To this end, we train by minimizing a Cross-Entropy Loss.

\begin{figure}[t]
    \centering
    \includegraphics[width=\linewidth]{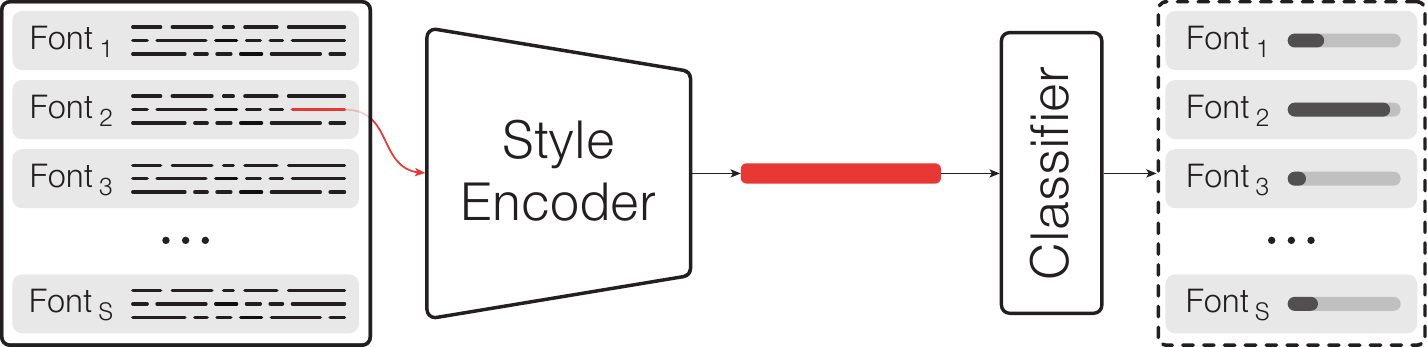}
    \caption{Overview of our pre-training approach. We train a single Style Encoder to classify the handwriting styles in the devised synthetic Font$^2$ dataset, and then we use it, frozen, to extract handwriting style representations that can be used in the four considered tasks: Writer Identification, Writer Retrieval, Writer Verification, and Writer Classification.}
    \label{fig:approach}
\end{figure}

Once pre-trained on the \dataname~dataset, we exploit the obtained handwriting style representation network for many writer-centric handwriting analysis tasks (see Figure~\ref{fig:approach}). Specifically, the style feature is obtained from the last convolutional block after the average pooling layer. In the following, we give the details on the specific strategies we apply to tackle each considered task.

\subsection{Writer Identification}
\begin{figure}[h]
\centering
\includegraphics[width=\linewidth]{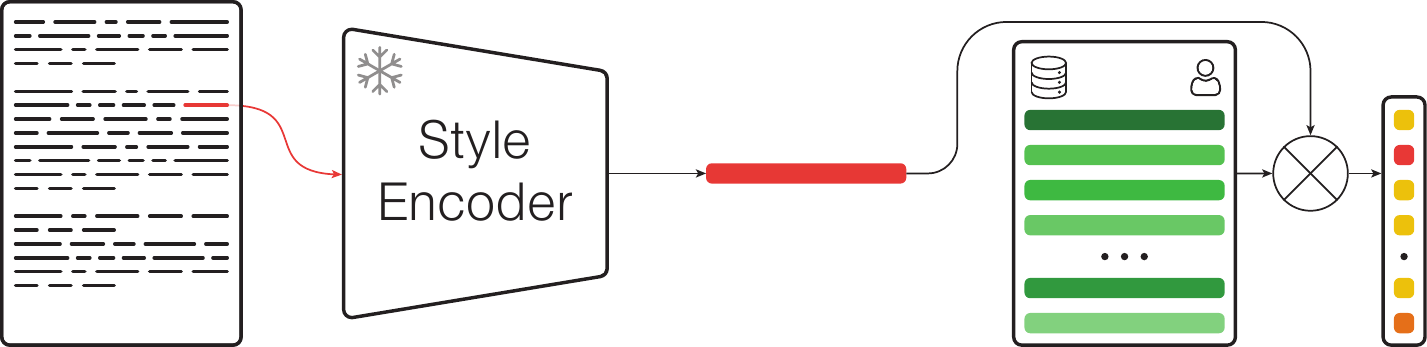}
\caption{Our Writer Identification pipeline. The red bars are the feature vectors extracted from words in the input document, which are compared with the task writers' feature vectors (in green) stored in a database of identifiable writers.}
\label{fig:identification}
\end{figure}

Writer Identification is performed at word level. Usually, the target datasets of real images for this task are split into training and test, each containing samples from each writer. 
For each writer in the training set, we compute an overall style representation by averaging all the style vectors of the words in the documents written by that writer. At inference time, we compare these style vectors with the style vector of each test word image via cosine similarity. The author of the test word image is then identified as the writer of the training documents whose style vector is the most similar to that of the test word. A schematic representation of our Writer Identification pipeline is reported in  Fig.~\ref{fig:identification}.

\subsection{Writer Retrieval}
\begin{figure}[h]
\centering
\includegraphics[width=\linewidth]{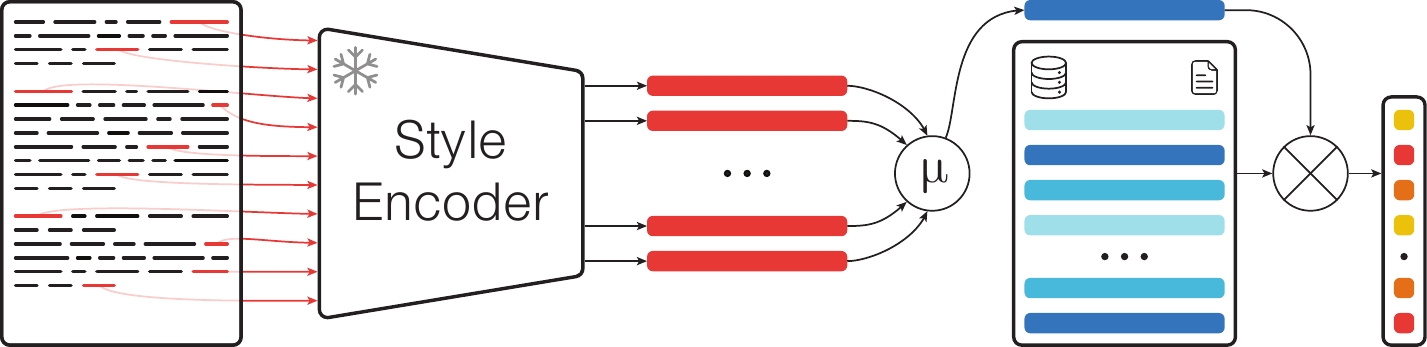}
\caption{Our Writer Retrieval pipeline. The blue bars are the feature vector of a specific document, obtained by averaging ($\mu$ symbol) the feature vectors of the words in the document. We compare the input document feature vector with those in the task documents database to retrieve all the documents from the same author of the input document.}
\label{fig:retrieval}
\end{figure}

Writer Retrieval is performed at document level. Also in this case, we represent each document by averaging the style vectors of its contained words. For retrieval, we compare the style vector of the query document with those of the other documents, which serve as database. The comparison is based on the Euclidean distance. As customary in the Writer Retrieval literature, we calculate the distance between the query and all the documents in the database, sort these latter, and output the closest $N$. Our Writer Retrieval pipeline is depicted in Fig.~\ref{fig:retrieval}.  An alternative strategy would be setting a threshold on the distance, \eg~based on a Precision-Recall curve, and outputting all the database documents whose distance from the query does not exceed the threshold.

\subsection{Writer Verification}
\begin{figure}[h]
\centering
\includegraphics[width=\linewidth]{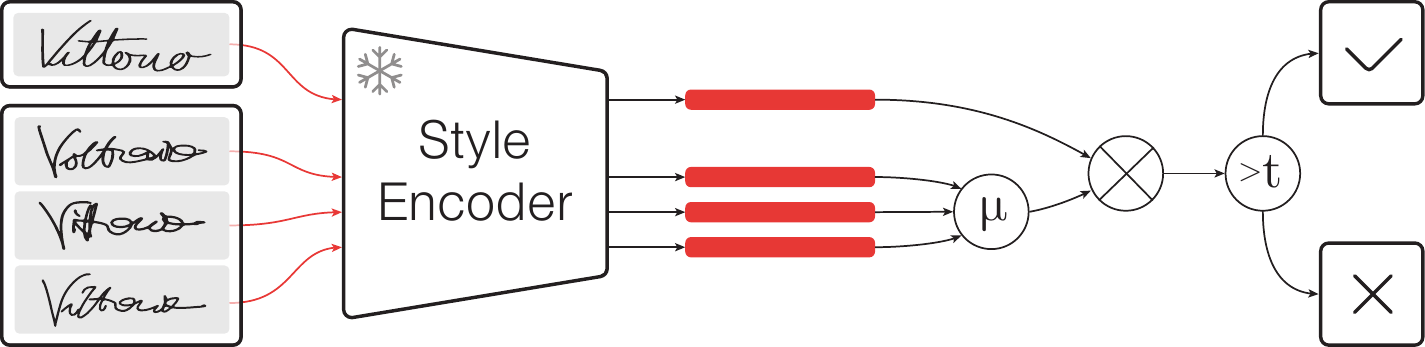}
\caption{Our Writer Verification pipeline, in the word-level setting (signature verification). We compute the distance between the mean feature of the real signature images ($\mu$ symbol) with the feature vector of the signature to verify. The signature is classified ad authentic if the distance between the feature vectors is above a given threshold, forged otherwise.}
\label{fig:verification}
\end{figure}

In this work, as a first Writer Verification use case, we consider writer independent signature verification, in which images of signatures must be compared to establish if both are genuine signatures of the same person (see Fig.~\ref{fig:verification}). Moreover, we consider Writer Verification at document level, where pairs of documents of multiple writers are compared to assess whether they have been written by the same person or not. In the signature verification case, we compare the style encoding vectors of query signature images with a template vector obtained by averaging the style vector of 5 genuine signature images. In the document-level case, we compare the style vectors of the entire page, obtained by averaging the vectors of its words. The comparison between vectors is performed via cosine similarity. Two signatures or documents are considered as belonging to the same writer if the similarity exceeds a certain threshold $t$, which can be determined, \eg~by observing a ROC curve.

\subsection{Writer classification}\label{ssec:approach_WC}
\begin{figure}[h]
\centering
\includegraphics[width=\linewidth]{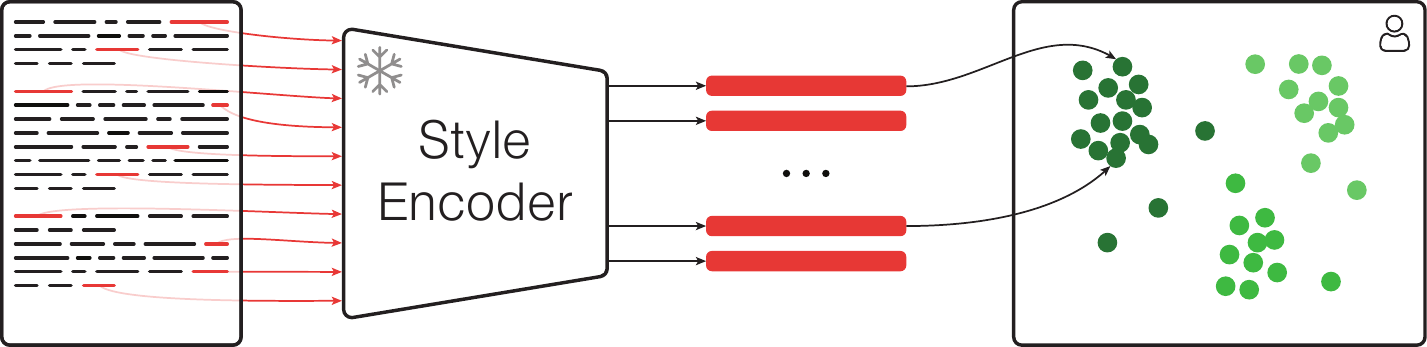}
\caption{Our Writer Classification pipeline. We cluster the feature vectors extracted from each document to group them based on the handwriting style, which should reflect the writers.}
\label{fig:classification}
\end{figure}

Writer Classification is performed as a clustering problem aimed at grouping all the images by each writer contained in the real image dataset of interest (Fig.~\ref{fig:classification}). 
In the case of datasets of word images, we consider the style vector of the single image, while for datasets of documents, we represent each document as the mean of the style vectors of its contained words. 
These vectors are then projected in a UMAP embedding~\cite{mcinnes2018umap} and input to an agglomerative clustering algorithm.  
Note that, to determine the number of writers in the dataset, which is usually not known a priori, we perform a grid-search on the hyperparameter $K$, \ie~number of desired clusters, and select the value that minimizes the silhouette score of the clustering.

\section{Experiments and Results}\label{sec:experiments}

We conduct our analysis on the role of synthetic data for handwriting processing tasks by considering the same architecture as backbone and varying the configuration of the \dataname~dataset used for pre-training. 
In particular, to evaluate the effects of the size of the pre-training dataset on the quality of the obtained handwriting style representation, we prepare variants of the \dataname~dataset by varying $S$ and setting it equal to 100, 500, 1000, 2000, and 5000, while maintaining all the available words. 
Once trained on the synthetic data, we test the model on real data.
Moreover, for reference, we also fine-tune the model pre-trained on the \dataname~dataset and train a baseline model from scratch on the target datasets of real images considered for the downstream tasks. 

\subsection{Implementation Details}
We train the network to classify the images in the \dataname~variants based on the font by optimizing the Cross-Entropy loss. We use Adam optimizer by setting the initial learning rate to $2 \times 10^-5$ with an exponential scheduling strategy with a decay factor of $10^{-1/90000}$. When training in all the dataset variants, we apply an early stopping strategy with patience 30 and batch size 32. Training time ranges from 1 to 10 days, depending on the \dataname~dataset split, on a single NVIDIA RTX 2080~Ti.
Due to the dataset size, every network converges before the end of the first epoch, \ie~before showing the network all the training samples, even with the smaller split. Therefore, for patience calculation, we consider 1000 training iterations (during which we feed the network $1000\cdot32$ samples) as an ``epoch''. 
After training, we discard the last fully connected layer, and the output of the Average pooling layer, which is a feature vector of length 512, is used for all downstream tasks. 

Regarding the fine-tuning experiments, we modify the last fully connected layer to fit the target dataset and train the whole network. After running some tests with different learning rates, we noticed that models trained with fewer classes on \dataname~perform better with a higher learning rate, while the more classes are used during pre-training, the lower the learning rate should be. Overall, a learning rate of $2 \times 10^-5$ is a good compromise, and we use it for all fine-tuning tests.

\subsection{Datasets}
For our experimental analysis on the downstream tasks considered, we use the following benchmark datasets. 
\tit{IAM~\cite{marti2002iam}} The IAM dataset contains 1452 documents handwritten in English by 657 writers. The number of pages per writer is highly variable, ranging from 1 to 10. The dataset comes with word-level segmentation information, which we use as input for our Style Encoder.
\tit{CVL~\cite{kleber2013cvl}} The CVL dataset features 1604 documents in English and German handwritten by 310 writers: 283 of them contributed 5 pages each, while the remaining 27 contributed 7 pages. Also for this dataset, word-level segmentation information is available.
\tit{CEDAR~\cite{kalera2004offline}} The CEDAR dataset for signature verification contains 2640 images of signatures from 55 different writers. Each writer provided 24 genuine signatures and several forged ones trying to imitate the other writers.

\subsection{Writer Identification}
We perform our analysis on the word-level Writer Identification task by considering the IAM and CVL datasets. The results are reported in Table~\ref{tab:wid} in terms of Top-1 and Top-5 Accuracy on the Writer Identification test sets of the two datasets as defined in~\cite{he2021gr}. As the goal of this study is to evaluate the effects of our proposed pre-training strategy, we mainly focus on different configurations of the synthetic dataset used. For completeness, we also report the results of some literature approaches, whose evaluation protocol (training/test splitting) is the closest to the one we use, although not perfectly comparable.

From Table \ref{tab:wid} emerges that specific knowledge of the target dataset is crucial for the identification task. In fact, when directly applying a model trained on a dataset that is not the target one, the identification performance is unsatisfactory. This applies both to literature approaches following this setting and to our pre-trained approach. However, pre-training on our proposed large synthetic dataset allows reaching better performance than pre-training on real ones. 
To explore the effect of fine-tuning, we consider the overall best-performing pre-trained variant, \ie~that trained on the 2000 split of~\dataname. After fine-tuning on the target dataset, the backbone pre-trained on \dataname~outperforms the baselines trained directly on the target datasets. These results demonstrate that the proposed pre-training strategy on \dataname, thanks to the variety of samples and the size of the dataset, allows obtaining strong and general style representations, which are useful in fine-tuning for the Writer Identification task.

\begin{table}[t]
\centering
\caption{Writer identification results. Bold indicates the best result, underlined the second-best among variants of our approach.\vspace{5pt}} 
\label{tab:wid}
\setlength{\tabcolsep}{.28em}
\resizebox{\linewidth}{!}{
\begin{tabular}{l c c c cc c cc}
\toprule
& & & & \multicolumn{2}{c}{\textbf{CVL}} & & \multicolumn{2}{c}{\textbf{IAM}}\\
\cmidrule{5-6} \cmidrule{8-9}
\textbf{Method} & & \textbf{Training data} & & \textbf{Top-1} & \textbf{Top-5} & & \textbf{Top-1} & \textbf{Top-5} \\
\midrule
\multirow{10}{*}{Ours}
& & 100 \dataname       & & 13.9 & 33.1 & & 13.6 & 31.9 \\
& & 500 \dataname       & & 21.4 & 43.5 & & 22.1 & 43.8 \\
& & 1000 \dataname      & & 25.8 & 49.0 & & 26.5 & 49.2 \\
& & 2000 \dataname      & & 27.4 & 51.2 & & 28.0 & 50.9 \\
& & 5000 \dataname      & & 25.1 & 48.1 & & 26.2 & 48.1 \\
& & 10400 \dataname     & & 25.9 & 48.8 & & 25.9 & 47.2 \\
\cmidrule{3-9}
& & CVL                 & & \underline{74.3} & \underline{92.7} & & 12.9 & 34.2 \\
& & IAM                 & & 15.9 & 36.2 & & \underline{69.3} & \underline{87.5} \\
\cmidrule{3-9}
& & 2000 \dataname+CVL      & & \textbf{82.0} & \textbf{94.9} & & 31.8 & 58.7 \\
& & 2000 \dataname+IAM      & & 31.8 & 57.1 & & \textbf{74.7} & \textbf{90.3} \\
\midrule
Siddiqui \& Vincent~\cite{siddiqi2010text} & & IAM & & 17.2 & 35.4 & & 16.9 & 33.0 \\
Brink \etal~\cite{brink2012writer}  & & - & & 23.8 & 46.7 & & 23.8 & 44.0 \\
He \& Shomaker~\cite{he2017beyond}  & & - & & 17.8 & 35.5 & & 20.9 & 37.4 \\
FragNet~\cite{he2020fragnet}        & & CVL/IAM & & 79.2 & 93.3 & & 72.2 & 88.0 \\
GR-RNN~\cite{he2021gr}              & & CVL/IAM & & 89.5 & 94.6 & & 83.3 & 94.0 \\
\bottomrule
\end{tabular}
}
\end{table}

\begin{table*}[t]
\centering
\caption{Writer retrieval results. Bold indicates the best result, underlined the second-best among variants of our approach.\vspace{5pt}} 
\label{tab:wrt}
\setlength{\tabcolsep}{.28em}
\resizebox{0.9\linewidth}{!}{
\begin{tabular}{l c c c cccccc c ccc}
\toprule
& & & & \multicolumn{6}{c}{\textbf{CVL}} & & \multicolumn{3}{c}{\textbf{IAM}}\\
\cmidrule{5-10} \cmidrule{12-14}
\textbf{Method} & & \textbf{Training data} & & \textbf{Top-1} & \textbf{STop-5} & \textbf{STop-10} & \textbf{HTop-2} & \textbf{HTop-3} & \textbf{HTop-4} & & \textbf{Top-1} & \textbf{STop-5} & \textbf{STop-10}\\
\midrule
\multirow{10}{*}{Ours}
& & 100 \dataname       & & 91.6 & 97.1 & 98.5 & 79.9 & 60.8 & 37.5 & & 91.0 & 95.3 & 96.6 \\
& & 500 \dataname       & & 96.8 & 98.9 & 99.1 & 92.4 & 80.6 & 56.1 & & 94.5 & 97.3 & \underline{97.6} \\
& & 1000 \dataname      & & 97.7 & 98.9 & 99.2 & 94.4 & 84.8 & 66.5 & & 95.4 & 97.3 & \textbf{97.7} \\
& & 2000 \dataname      & & 97.9 & 99.0 & 99.2 & 95.0 & 88.0 & 69.9 & & \underline{95.9} & \underline{97.4} & 97.5 \\
& & 5000 \dataname      & & 98.2 & \underline{99.2} & \textbf{99.4} & 95.4 & 89.5 & 74.9 & & 95.8 & 97.3 & \textbf{97.7} \\
& & 10400 \dataname     & & 98.4 & 98.9 & \underline{99.3} & \underline{95.9} & 90.3 & 76.8 & & \textbf{96.0} & \textbf{97.5} & \underline{97.6} \\
\cmidrule{3-14}
& & CVL                 & & \textbf{99.1} & 99.1 & \underline{99.3} & \textbf{97.5} & \underline{94.8} & \underline{88.7} & & 84.1 & 92.7 & 94.5 \\
& & IAM                 & & 82.2 & 93.8 & 96.1 & 63.2 & 44.5 & 26.0 & & 92.6 & 96.4 & 97.1 \\
\cmidrule{3-14}
& & 10400 \dataname+IAM & & 93.1 & 95.9 & 96.7 & 80.9 & 63.2 & 34.2 & & 95.8 & 97.0 & 97.4 \\
& & 10400 \dataname+CVL & & \underline{99.0} & \textbf{99.3} & \underline{99.3} & \textbf{97.5} & \textbf{95.1} & \textbf{89.9} & & 93.7 & 95.6 & 96.1 \\
\midrule
Fiel \& Sablatnig~\cite{fiel2012writer} & & TrigraphSlant    & & - & - & - & - & - & - & & 93.1 & 96.7 & 97.6 \\
Fiel \& Sablatnig~\cite{fiel2015writer} & & IAM              & & 98.9 & 99.3 & 99.5 & 97.6 & 93.3 & 79.9 & & - & - & - \\
Bouibed \etal~\cite{bouibed2020multiple} & & ICDAR17         & & -    & - & - & 99.6 & 99.0 & 98.0 & & - & - & - \\
Rasoulzadeh \& Baba~\cite{rasoulzadeh2022writer} & & ICDAR13 & & 99.2 & - & - & 98.9 & 98.0 & - & & - & - & - \\
\bottomrule
\end{tabular}
}
\end{table*}

\subsection{Writer Retrieval}
In the Writer Retrieval experiments, we exploit both the IAM and CVL datasets. For the IAM dataset, we include in the analysis only those writers with at least two pages (301). Note that, for these experiments, each document serves as query in a leave-on-out scheme. 
The results are reported in Table~\ref{tab:wrt} in terms of Top-1 Accuracy, Soft Top-5 and Soft Top-10 Accuracy (STop-5 and STop-10, respectively), and Hard Top-2, Hard Top-3, and Hard Top-4 Accuracy (HTop-2, HTop-3, and HTop-4, respectively). Note that for calculating the Soft Top-$N$ scores, we consider as a valid output the cases in which at least one correct sample is retrieved among the first $N$, while for Hard Top-$N$ scores, valid outputs are the cases in which all the first $N$ retrieved samples are correct. Also in this case, we report the results of the literature approaches that do not use the target dataset for training. The considered approaches resort to real publicly available datasets, such as TrigraphSalnt~\cite{brink2011towards}, ICDAR13~\cite{louloudis2013icdar}, and ICDAR17~\cite{fiel2017icdar2017}.

From Table~\ref{tab:wrt} emerges that pre-train on our synthetic dataset without fine-tuning allows achieving competitive results in the Writer Retrieval task. The model trained on the 10400 \dataname~split achieves results very close to the baseline trained on CVL and, at the same time, outperforms the baseline trained on IAM. However, fine-tuning the model on the target dataset improves the performance on CVL, but on the other hand, fine-tuning on IAM achieves the opposite effect by degrading the performance.
The higher Hard Top-$N$ accuracy obtained by the literature approaches compared to our model can be imputed to their use of ad hoc techniques to increase the retrieval results (\eg ~reranking). Nevertheless, by resorting to the proposed pre-training strategy, we are able to reach competitive performance despite the simplicity of the pipeline adopted.

\subsection{Writer Verification}
For Writer Verification at word level (\ie~signature verification), we use the CEDAR dataset, while for the document-level setup, we consider the IAM and CVL datasets (see Table~\ref{tab:wver}). Also in this case, for the IAM dataset, we include in the analysis only those writers with at least two pages.
We report the performance in terms of Equal Error Rate (ERR), which is the best trade-off achievable between the False Acceptance Rate and the False Rejection Rate, \ie~the probability of a forged document being recognized as genuine and the probability of a genuine document being recognized as forged. 
For completeness, we report the results of some task-specific literature proposals deemed more comparable to our approach for not using the target dataset in training. Indeed, these are trained on public datasets such as GPDS Synthetic~\cite{ferrer2013synthetic}, MCYT75~\cite{galbally2015line}, BiosecurIDSONOF~\cite{liwicki2011signature}, and SigComp11~\cite{liwicki2011signature}. 

The results obtained show that the generalization capabilities of the model increase when trained on the \dataname~dataset. In fact, for the document-level tests on the IAM and CVL datasets, the verification performance of the models pre-trained on \dataname~only is comparable to models trained directly on the target dataset. 
On the word-level CEDAR dataset, on the other hand, the models trained on \dataname~struggle to achieve good results, probably due to the difference between our synthetic dataset and the target domain.  
Moreover, the model trained only on a subset of CEDAR performs poorly due to the dataset size.
From the fine-tuning experiments, which we carry out considering the best-performing pre-trained model (the one using the 5000 \dataname~split), it emerges that fine-tuning on IAM and CEDAR is particularly convenient for significantly improving the performance while the benefit of fine-tuning on CVL is more evident when testing on other datasets than on CVL itself. 

\begin{table}[t]
\centering
\caption{Writer verification results in terms of ERR. Bold indicates the best result, underlined the second-best among variants of our approach.\vspace{5pt}} 
\label{tab:wver}
\setlength{\tabcolsep}{.28em}
\resizebox{0.9\linewidth}{!}{
\begin{tabular}{l c c c ccc}
\toprule
\textbf{Method} & & \textbf{Training data} & & \textbf{CVL} & \textbf{IAM} & \textbf{CEDAR} \\
\midrule
\multirow{12}{*}{Ours}
& & 100 \dataname       & & 4.0 & 4.2 & 18.0 \\
& & 500 \dataname       & & \underline{2.2} & 4.1 & 15.9 \\
& & 1000 \dataname      & & 2.7 & 3.8 & 14.7 \\
& & 2000 \dataname      & & 2.4 & 4.2 & 13.8 \\
& & 5000 \dataname      & & 2.3 & 3.6 & 13.1 \\
& & 10400 \dataname     & & 2.4 & 3.7 & 13.9 \\
\cmidrule{3-7}
& & CVL                 & & \textbf{1.9} & 8.5 & 12.0 \\
& & IAM                 & & 6.4 & \underline{3.5} & 19.4 \\
& & CEDAR               & & 36.4 & 29.0 & 24.8 \\
\cmidrule{3-7}
& & 5000 \dataname+CVL  & & \textbf{1.9} & 5.2 & \underline{11.5} \\
& & 5000 \dataname+IAM  & & 4.0 & \textbf{1.8} & 15.4 \\
& & 5000 \dataname+CEDAR    & & 16.2 & 9.4 & \textbf{10.4} \\
\midrule
Wan \& Zou~\cite{wan2021learning} & & GPDS Synthetic & & - & - & 8.1 \\
Zhu \etal~\cite{zhu2020point} & &  Multiple\footnotemark[1] & & - & - & 9.1 \\
\bottomrule
\addlinespace[0.1cm]
\multicolumn{7}{l}{\footnotesize 1 -  MCYT75 + BiosecurIDSONOF + SigComp11}
\end{tabular}
}
\end{table}

\subsection{Writer Classification}
In the case of Writer Classification, we consider all the documents in the IAM and CVL datasets and all the genuine signatures in the CEDAR dataset since their authorship corresponds to the author label provided as ground truth. As explained in Section~\ref{ssec:approach_WC}, the Writer Classification task is cast as a clustering one, where the number of clusters, $K$, corresponds to the number of writers in the target dataset. In case this number is not known, we indicate our best estimate of $K$ as $K'$. In the scenario in which the actual number of writers is known, we indicate this number as $K^*$ and use it for clustering. 

We first perform an experiment simulating the case in which the number of writers that contributed to a collection of documents is unknown. In this scenario, the performance is expressed in terms of the difference between the real and the best-estimated number of writers, \ie~$\Delta K = K^* - K'$. The results are reported in Table~\ref{tab:wcl}. 
For this scenario, the most accurate variant is the one pre-trained on the complete \dataname~dataset, whose $\Delta K$ is on average smaller than the other variants on the target datasets. This value is further reduced after fine-tuning on the target datasets. From these results emerges that pre-training on a sufficient amount of data leads to an informative representation of the writer's style, which is also enforced by the high $\Delta K$ obtained when pre-training on a small dataset such as CEDAR.

Moreover, we report the performance achievable when $K = K^*$, \ie~the number of writers is known a priori. Also the results of this study are reported in Table~\ref{tab:wcl}. It can be noticed that on datasets in which all the featured writers are represented by a sufficient number of samples (5/7 in CVL, 24 in CEDAR), the fine-tuned variant of the proposed approach brings limited benefit with respect to the simply pre-trained counterpart. However, in the more challenging IAM dataset, the effect of fine-tuning is more evident. Note that, in general, the poor performance of all the considered variants on the IAM dataset can be imputed to the fact that for more than half of the writers in the dataset, there is only one document available, making clustering particularly difficult.

\begin{table}[t]
\centering
\caption{Writer classification results. K$^*$-Sil. is the Silhouette score obtained by clustering the authors in a number of clusters that is equal to the real number of authors in the dataset. Bold indicates the best result, underlined the second-best.\vspace{5pt}} 
\label{tab:wcl}
\setlength{\tabcolsep}{.28em}
\resizebox{\linewidth}{!}{
\begin{tabular}{c c cc c cc c cc}
\toprule
\textbf{Training data} & & \multicolumn{2}{c}{\textbf{CVL}} & &  \multicolumn{2}{c}{\textbf{IAM}} & &  \multicolumn{2}{c}{\textbf{CEDAR G.}}\\
\cmidrule{3-4} \cmidrule{6-7} \cmidrule{9-10}
& & \textbf{$K^*$-Sil.} & \textbf{$\Delta K$} & &  \textbf{$K^*$-Sil.} & \textbf{$\Delta K$} & &  \textbf{$K^*$-Sil.} & \textbf{$\Delta K$}\\
\midrule
100 \dataname       & & 0.97 & +2 
                    & & 0.54 & +415
                    & & 0.68 & -53\\
500 \dataname       & & \underline{0.98} & \textbf{0} 
                    & & 0.57 & +372 
                    & & 0.84 & +70\\
1000 \dataname      & & \textbf{0.99} & +1 
                    & & 0.56 & +361 
                    & & 0.84 & +15\\
2000 \dataname      & & \textbf{0.99} & -2 
                    & & 0.57 & +353 
                    & & 0.90 & +10\\
5000 \dataname      & & \textbf{0.99} & \underline{-1} 
                    & & 0.58 & +392 
                    & & \underline{0.91} & +6 \\
10400 \dataname     & & \underline{0.98} & \underline{-1} 
                    & & 0.56 & \underline{+314} 
                    & & 0.88 & \underline{+3} \\
\midrule
CVL                 & & \textbf{0.99} & \underline{-1} 
                    & & 0.54 & +409 
                    & & 0.70 & +4 \\
IAM                 & & 0.92 & \underline{-1} 
                    & & 0.60 & +388 
                    & & 0.56 & +5 \\
CEDAR               & & 0.53 & +295 
                    & & 0.57 & +650 
                    & & 0.51 & +53 \\
\midrule
10400 \dataname+CVL    & & \textbf{0.99} & +3 
                       & & \underline{0.69} & +339 
                       & & 0.87 & -3 \\
10400 \dataname+IAM    & & 0.96 & -6          
                       & & \textbf{0.71} & \textbf{+303} 
                       & & 0.86 & -15 \\
10400 \dataname+CEDAR  & & 0.93 & -20         
                       & & 0.66 & +655 
                       & & \textbf{0.92} & \textbf{-1} \\
\bottomrule
\end{tabular}
}
\end{table}

\section{Conclusion}\label{sec:conclusion}
In this work, we have presented a fully supervised pre-training protocol on carefully designed synthetic data to obtain a robust and expressive handwriting Style Encoder. The effectiveness of our proposed approach has been analyzed by varying the size of the pre-training dataset and by exploiting the obtained representations on four downstream handwriting analysis tasks on real images from benchmark datasets, namely Writer Identification, Writer Retrieval, Writer Verification, and Writer Classification. 

The experimental results demonstrate the suitability of the proposed strategy and show that the obtained style representation is overall robust. Nonetheless, some guidelines can be traced to apply the most suitable configuration of our proposed approach, which are discussed in the following.
Some handwriting analysis tasks, such as Writer Identification and Writer Verification, strongly depend on the specific data to manage. For this reason, in these tasks, just pre-training offers limited advantages. Nevertheless, after fine-tuning on the target dataset, we gain a consistent performance improvement over training on the target data only. When fine-tuning for these dataset-depending tasks, using a Style Encoder pre-trained on smaller versions of \dataname~(\eg~on the variant with $S=2000$ or $S=5000$) leads to the best performance while also requiring less pre-training time. Note that the amount of samples in those variants is enough to train a relatively simple network (as the ResNet-18 model that we use) capable of extracting robust features.
On the other hand, for other tasks, such as Writer Retrieval and Writer Classification, the knowledge about the target dataset is less important than the robustness of the data representation. On those tasks, pre-training on the bigger, more varied versions of \dataname~allows obtaining such a strong representation, thus leading to the best performance even without fine-tuning on the task-specific target dataset.

\section*{Acknowledgement}
This work was supported by the ``AI for Digital Humanities'' project (Pratica Sime n.2018.0390), funded by ``Fondazione di Modena''.

\bibliography{main.bib}

\end{document}